# Dynamic Chain-of-Thought: Towards Adaptive Deep Reasoning

Libo Wang
Nicolaus Copernicus University
Jurija Gagarina 11, 87-100 Toruń, Poland
326360@o365.stud.umk.pl
UCSI University
Taman Connaught, 56000 Kuala Lumpur, Wilayah Persekutuan Kuala Lumpur, Malaysia
1002265630@ucsi.university.edu.my

*Abstract*— **To reduce the cost and consumption of computing resources caused by computational redundancy and delayed reward assignment in long CoT, this research proposes the dynamic chain-of-thought (D-CoT) with adaptive reasoning time and steps. The researcher used simulation experiment to simulate the integration of D-CoT through Python 3.13 IDLE combined with a Python simulator based on GPTs. At the same time, the researcher used DeepSeek R1 as a control group to test and compare the performance of the D-CoT simulator in processing MIT OpenCourseWare's linear algebra exam questions. Experimental results show that D-CoT is better than DeepSeek R1 based on long CoT in three indicators: reasoning time, CoT length (reasoning steps) and token count, which achieves a significant reduction in computing resource consumption. In addition, this research has potential value in deep reasoning optimization that is used as a reference for future dynamic deep reasoning frameworks.**

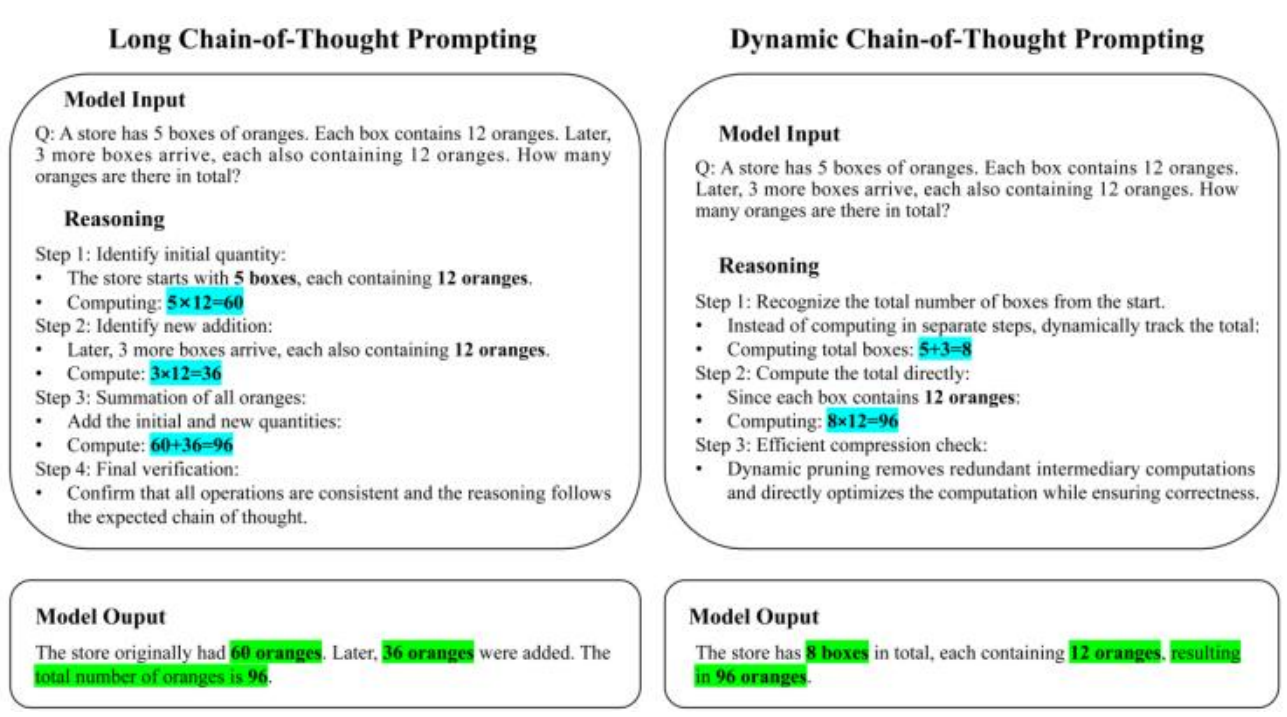

Figure 1 - Comparison of the reasoning process of long CoT and dynamic CoT via prompting.

## I. INTRODUCTION

As an emergent capability of large language models (LLMs) with huge parameter sizes in the reasoning process, chain-of-thought (CoT) allocates additional computing resources in a way that facilitates the gradual decomposition of tasks, which is particularly prominent in context learning (Wei et al. al., 2022). It follows that chain-of-thought provides techniques for gradually unfolding intermediate reasoning steps to enhance LLMs' ability to handle complex problems by disassembling them into coherent sub-steps (Feng et al., 2024). However, affected by context, prompt word design, and model learning bias, CoT has unfaithful interpretations, which leads to deviations between the reasoning process and the actual decision-making mechanism (Turpin et al., 2024). In addition, while CoT enhances the reasoning capabilities of LLMs, it also brings concerns about rising costs by prolonging the reasoning steps and improving quality (Jin et al., 2024). In response to the above-mentioned shortcomings of traditional CoT, more and more cutting-edge LLMs apply long CoT technology to demonstrate excellent reasoning capabilities when processing complex tasks (Chen et al., 2024; Wang et al., 2024).

Long chain-of-thought (long CoT) aims to promote the model to have self-reflection and adaptive refinement in multi-step complex scene reasoning through hierarchical reasoning and stepwise verification to ensure accuracy and consistency (Wang et al., 2024). For few-shot CoT, the accuracy of LLMs output is linearly related to the number of reasoning steps, which means that the longer the number of steps, the more accurate the response, while reducing the CoT length will significantly reduce the response accuracy (Jin et al., 2024). Jin et al. found that even if errors occur in the intermediate steps of long CoT during the reasoning process, maintaining the necessary reasoning length will produce a high-accuracy response.

In the application of OpenAI's o1 model, long CoT is combined with reinforcement fine-tuning (RFT) to further optimize LLMs' understanding and memory capabilities of multi-level reasoning processes (Huang et al., 2024; Zhang et al., 2024). In view of the shortcomings of LLMs in intermediate reasoning and adaptive learning capabilities, DeepSeek-R1 introduces large-scale pure reinforcement learning training to reduce reliance on supervised fine-tuning (SFT) (Guo et al., 2025). During training, in order to avoid instability in the initial cold start phase due to no longer relying on large amounts of labeled data, it constructs long CoT data and uses specific collection and processing methods to guide deeper reflection. and verification (Chen et al., 2024; Liu et al., 2024).

However, long CoT requires a large number of intermediate steps in the reasoning practice process, its inherent computational redundancy and delayed feedback problems significantly increase the reasoning cost and consumption of computing resources, which is directly reflected in the exponential growth of reasoning time and steps. In order to improve accuracy, a large number of lengthy reasoning steps do not directly contribute to the final answer but only serve as an auxiliary process, resulting in the accumulation of computational overhead (Dai et al., 2024). These phenomena are often reflected in users' actual applications, especially LLMs with deep reasoning capabilities such as o3 min-high or DeepSeek R1. For example, when users use DeepSeek R1 to perform difficult and complex tasks, the number of reasoning steps increases significantly and the system response delay increases significantly. Unstable reasoning fluctuations cause low efficiency, and even the response "The server is busy. Please try again later" appears frequently due to server resource saturation.

The deep gap lies in the fact that long CoT essentially runs on a static reasoning framework, making it difficult to flexibly adjust the number of reasoning steps according to the difficulty of different problems. The lack of dynamic adaptation mechanism usually treats the reasoning process as a linear expansion, which is reflected in the inability to adjust the length of the thinking chain according to the complexity of different tasks and environmental feedback. For example, when DeepSeek R1 faces deep reasoning on difficult tasks, even if some reasoning steps have minimal impact on the final decision, it still cannot actively compress or ignore redundant steps. This design flaw makes it difficult for the system to adaptively allocate computing resources, and the accumulation of inefficient reasoning causes nonlinear growth in reasoning time. In addition, the insufficient coupling between long CoT generation and RL reward mechanism is one of gaps about huge computational overhead. This technology is designed to enhance transparency and explainability, but is not tightly integrated with RL goals, which results in the lack of an effective credit allocation mechanism between reward signals during reasoning. Even though traditional RL relies on immediate or deferred rewards to adjust strategies, the value of a step in a long CoT is usually only determined when rewards are finally obtained.

## II. PROPOSED MODULE & ALGORITHMS

In view of gaps, this research proposes dynamic chain-of-thought (D-CoT) to implement a state compression mechanism with adaptive reasoning steps to reduce computational redundancy. Fig. 1 shows the comparison of the reasoning process between long CoT and dynamic CoT under the same prompt.

### *2.1. Dynamic Chain-of-Thought Framework*

Dynamic chain-of-thought (D-CoT) is an LLMs reasoning framework with adaptive reasoning capabilities that reduces the consumption of cost computing resources by real-time adjustment of chain length (reasoning steps) and reasoning time. Compared with the fixed and linear expansion of reasoning steps of traditional long CoT, D-CoT dynamically adjust the length of CoT in real time, select key steps after rating different tasks. Its specific internal structure is shown in Fig. 2.

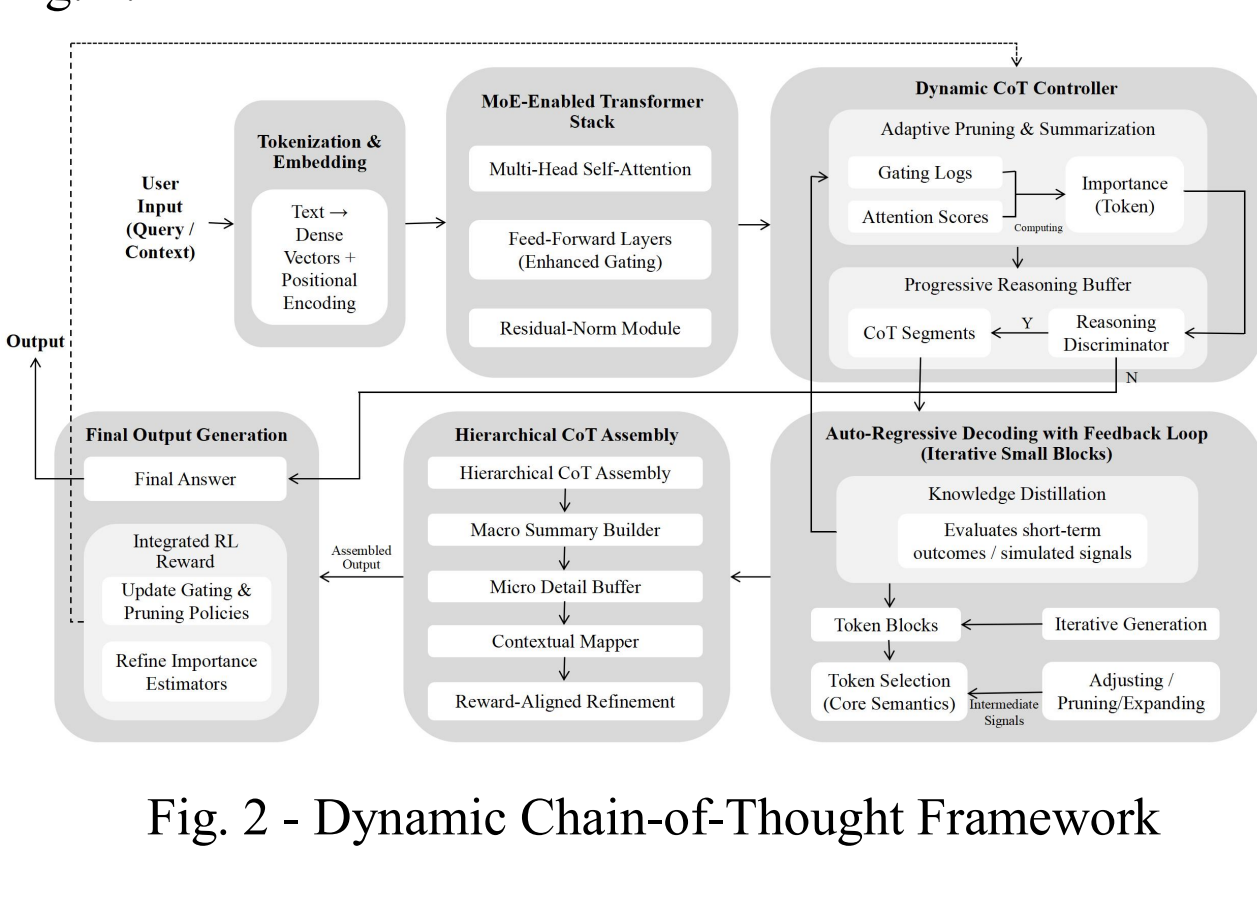

Fig. 2 - Dynamic Chain-of-Thought Framework

The D-CoT leverages the core principles of the hierarchical adaptive reinforcement learning to adjust the steps and information weights in the deep reasoning process to minimize computational redundancy and optimize the decision path. It introduces an importance-driven pruning strategy in the process of auto-regressive decoding, combines the partial reward estimator to instantly evaluate the effectiveness of the reasoning block, and decides whether to expand or delete the reasoning steps through adaptive thresholding. In addition, D-CoT constructs a multi-level reasoning structure through macro summary and micro detail buffer to ensure the optimal transmission of information flow at different reasoning scales. In contrast, it not only reduces the cumulative computational burden of long CoTs, but also improves the adaptability of reasoning, forming an efficient reasoning framework with feedback adjustment capabilities. The following is a detailed display of the HARO algorithm:

#### *2.1.1. Hierarchical Adaptive Reward Optimization*

The operating mechanism of the HARO (Hierarchical Adaptive Reward Optimization) algorithm is based on hierarchical reward allocation and adaptive reasoning adjustment. The algorithm uses the partial reward estimator to instantly evaluate the contribution of decisions at different levels of the deep reasoning process, and uses adaptive thresholding to dynamically correct the weight of the reasoning step (CoT length) to ensure the priority delivery of high-value information. In addition, this algorithm combines an importance-driven pruning strategy to instantly filter inefficient reasoning paths to reduce redundant computing overhead and improve overall reasoning efficiency.

- Token Importance Evaluation

Each reasoning step $c_i$ is assigned an importance score:

$$I(c_i) = \alpha * A(c_i) + (1 - \alpha) * \text{GatingScore}(c_i)$$

where $A(c_i)$ is the dominance estimate derived from RL and GatingScore reflects the token-level contribution.

- Dynamic Adaptive Pruning Thresholding

It introduces a self-adjusting threshold $\tau_t$ based on historical success rates:

$$\tau_t = \gamma \cdot \tau_{t-1} + (1-\gamma) \cdot \frac{1}{N} \sum_{j=1}^{N} 1[I(c_j)] > \tau_{t-1}]$$

where $\tau_t$ represents the updated threshold at time step $t$; $\gamma$ is an attenuation factor used to control the retention of historical information; $1[I(c_j)>\tau_{t-1}]$ tracks whether past tokens have exceeded a previous threshold.

- Progressive Reasoning Buffer (Adaptive Selection)

Dynamic adjustments in CoT segments are stored in buffers:

$$C_t = C_{t-1} + \text{argmax}_{ci} (I(c_i) - \tau_t)$$

where steps below $\tau_t$ are discarded unless they contribute significantly to global coherence; $C_t$ represents the CoT state at time $t$.

- Reward Optimization and Auto-Regressive Feedback

HARO uses reward gradients to iteratively optimize token selection by core semantics:

$$\nabla_{\theta} J = \boldsymbol{E}\,[R_{\text{sem}}(C) + \lambda R_{\text{struct}}(C)) \nabla_{\theta} \log \pi_{\theta}(C)]$$

where $\boldsymbol{E}$ represents expectation value; $R$ represents reward function; $\theta$ represents model parameters; $R_{\text{sem}}(C)$ represents a semantic reward function; $R_{\text{struct}}(C)$ represents a structural function; λ is a weighting hyperparameter balancing semantic fidelity and structural efficiency; $\pi_{\theta}(C)$ represents the policy for selecting tokens; $\nabla_{\theta} \log \pi_{\theta}(C)$ is the optimization through policy gradients.

Notably, reward alignment and policy adjustment are inspired by PPO (Proximal Policy Optimization). PPO is committed to truncating the clipped objective function restriction policy and updating the amplitude policy update range to ensure training stability (Schulman et al., 2017). HARO further introduces adaptive thresholding, allowing reward signals to dynamically adapt according to the reasoning steps to improve the selectivity of the optimal decision trajectory. In addition, HARO draws on advantage estimation, dynamically filters high-value reasoning through hierarchical feedback mechanism, reduces low-reward expansion, and thereby reasons redundancy.

### *2.2. Detailed Framework Composition*

The D-CoT framework consists of six key parts. First, it converts user input into dense vectors through tokenization and embedding. Then, the MoE-enabled transformer stack is applied to the multi-head self-attention mechanism to enhance semantic expression (Dai et al., 2024). Dynamic CoT Controller performs reasoning step screening through adaptive pruning and attention scores. Subsequently, auto-regressive decoding uses partial reward estimation for mark selection and incremental generation. Finally, the hierarchical CoT assembly integrates the deep reasoning process and uses reward-aligned refinement to optimize the final output.

#### *2.2.1 Tokenization & Embedding*

As the first part of D-CoT, it is responsible for converting natural language input into a dense vector representation that can be processed by the model (Tunstall et al., 2022). This process first decomposes the text into subword units through tokenization, maps it to a high-dimensional space through an embedding layer to capture semantic information and contextual dependencies, and combines it with positional encoding to provide sequence order information (Vaswani, 2017). The following are the supported algorithms in the workflow:

- Tokenization Process

$$\mathbf{T} = \text{Tokenizer}\,(\mathbf{Q})$$

- Embedding with Positional Encoding

$$\mathbf{X} = E(\mathbf{T}) + P,$$

- Seamless Transition to MoE Stack

$$\mathbf{X} \longrightarrow \text{MoE-Enabled Transformer}$$

where converting user query **Q** into a token sequence **T**, this algorithm converts **Q** into discrete labeled units; Mapping tokens into dense embeddings while integrating positional encoding; **X** represents the final embedding representation that contains the vectorized representation and position encoding; $E$(T) represents the embedding function that converts mark tokens into corresponding embedding representations; $P$ represents position encoding.

#### *2.2.2 MoE-Enabled Transformer Stack*

The MoE-enabled transformer stack uses a mixture of experts (MoE) mechanism to enhance selective computing capabilities through multi-head self-attention to ensure efficient acquisition of key information (Liu et al., 2024). The feed-forward layers combined with enhanced gating perform feature transformation based on dynamically selected experts to maximize reasoning efficiency (Guo et al., 2025). The residual-norm module provides gradient stability and reduces signal attenuation to ensure smooth flow of information in deep structures. The researcher demonstrated its workflow algorithm as follow.

- Expert Selection

$$\alpha_{t,e} = \text{Router}\,(u_t, e),\ \text{E}_{\text{active}} = \text{TopK}\{\alpha_{t,e} \mid e = 1,\ldots, N\}$$

- Expert Outputs

$$\mathrm{h}_t' = \sum_{e \in E_{t,\text{active}}} \alpha_{t,e} \cdot f_e(\mathrm{u}_t), \mathrm{u}_t \in \Re$$

- Importance Score

$$I_t = \gamma\,(\sum_{e \in \mathrm{E}_{t,\text{active}}} \alpha_{\mathrm{t,e}}) + (1-\gamma)\,\beta_t$$

After expert assignment of input vectors through MoE, the refined semantic information is passed to the dynamic CoT controller for adaptive pruning and dynamic summarization to reduce reasoning redundancy (Fig. 3).

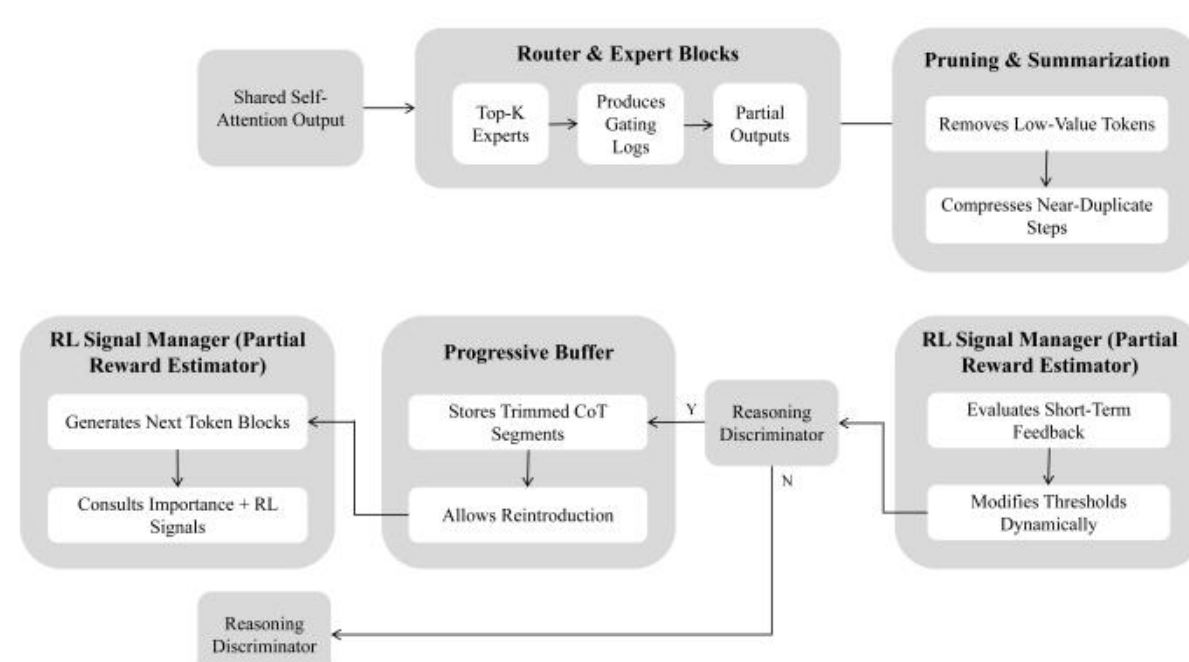

Fig.3 - The connection between MoE-enabled transformer stack and dynamic CoT controller.

- Adaptive Thresholding for Pruning and Summarization

$$\tau_{\text{dyn}}\,(r_t) = \tau_0 + \eta(r_t - \bar{r})$$

- Hierarchical Decoding & Assembly

$$y_{t+1} = \text{Assemble}\,(B_T, \{\text{macro, micro}\})$$

where $\tau_{\text{dyn}}(r_t)$ is dynamic threshold; $r_t$ is partial reward; $\tau_0$ is a base threshold, $\bar{r}$ is a running average reward, and $\eta$ is a scaling factor; $y_{t+1}$ represents the next generated token after decoding; $B_T$ represents the final buffer of CoT segments after iterative refinement; The macro-level summaries compress global information; micro-level expansions retain fine-grained details.

*2.2.3 Dynamic CoT Controller*

As the core of D-CoT's deep reasoning, the dynamic CoT controller is responsible for optimizing the length and content of CoT to reduce redundant calculations and the accumulation of unnecessary steps. The key technology is to dynamically adjust the reasoning steps to adapt the reasoning process to different types of task requirements, thereby improving the adaptability of LLMs in complex decision-making scenarios. Its operating mechanism calculates the importance score of the token based on gating logs and attention scores, which determines whether the inference step should be retained or compressed. Afterwards, the reasoning fragments are stored in the progressive reasoning buffer to ensure that key information is efficiently used in subsequent steps. Reasoning discriminator determines whether to start CoT reasoning through knowledge certainty evaluation (Y/N). The system calculates the confidence $P_{\text{fact}}(x)$ via FAISS/BM25, $C_{\text{comp}}(x)$ is not greater than 3 that is based on syntactic structure and computational cost evaluation. If both are below the threshold, they are directly output, otherwise they enter CoT segments for reasoning. The relevant supported algorithms are as follows:

- Adaptive Pruning & Summarization

$$I_t = \text{Importance}\ (t)\ \text{(combining gating + attention)}$$

$$\text{if } I_t < \tau_{\text{dyn}}(r_t),\ \text{prune token } t,\ \text{else, optionally summarize}$$

- Progressive Buffer Update

$$B_{t+1} = \text{B}_t \cup \{\text{Summarized } t \mid I_t \geq \tau_{\text{dyn}}(r_t)\}$$

- Partial Reward

$$r_t = \text{RLFeedback}\ (t),\ \tau_{\text{dyn}}(r_t) = \tau_0 + \eta \bullet (r_t - \bar{r})$$

- Adaptive Token Expansion and Pruning

$$\text{z}_{t+1} = \text{Adjust}\ (\text{Generate}\ (\text{SelectTokens}\ (B_t, \theta_t, r_t), B_t, \theta_t), \text{r}_{t+1})$$

- Reasoning Discriminator

$$y_{\text{out}} = \begin{cases} \text{OutputAnswer}(x), \text{if } P_{\text{fact}}(\text{x}) \geq 0.85 \text{ and } C_{\text{comp}}(x) \leq 3 \\ \text{CoT Segments}(x), \text{otherwise} \end{cases}$$

- Reward-Guided CoT Assembly

$B_t$ = AssembleCoT($B_t$, $\{z_t\}_{t=-T}$, $\text{r}_{t+1}$ = RewardUpdate ($r_t$, DecodeOut ($\text{z}_{t+1}$)))

where $I_t$ is importance score of token t (gating + attention); $\tau_{\text{dyn}}$ represents dynamic threshold based on reward $r_t$; $\tau_0$ is base pruning threshold; $\eta$ is scaling factor for threshold adjustment; $r_t$ is RL feedback at step $t$ / partial reward signal; $\text{r}_{t+1}$ represents the updated reward at step $t$+1; $\bar{r}$ represents running average reward; $B_t$ represents token buffer at step $t$ / the final structured CoT assembly after multiple iteration steps; $B_{t+1}$ represents updated token buffer; $t$ represents the current reasoning step in the iterative decoding process; Summarized $t$ is compressed token representation; RLFeedback($t$) represents RL-based reward function for token $t$; $z_t$ represents tokens at time step $t$ / the generated reasoning tokens at step $t$; $\theta_t$ represents hidden parameters of the decoder at $t$; SelectTokens() represents token selection of of core semantics; Generate(·) is based on generation of new tokens; Adjust(·) expands or prunes tokens (adaptive pruning or expansion decision) based on rewards and gating logs; $T$ is The total number of CoT reasoning steps; RewardUpdate() updates the reward based on the previous reward and newly decoded token sequence; DecodeOut($\text{z}_{t+1}$) is from the iterative reasoning process at step $t$+1; AssembleCoT() structures the final CoT by integrating generated reasoning tokens and their associated reward values.

The processed CoT fragment is passed to auto-regressive decoding with feedback loop (Fig.4). It evaluates the effectiveness of each fragment based on the partial reward estimator and optimizes the reasoning strategy through dynamic adjustment, which reduces computational overhead while maintaining high accuracy output.

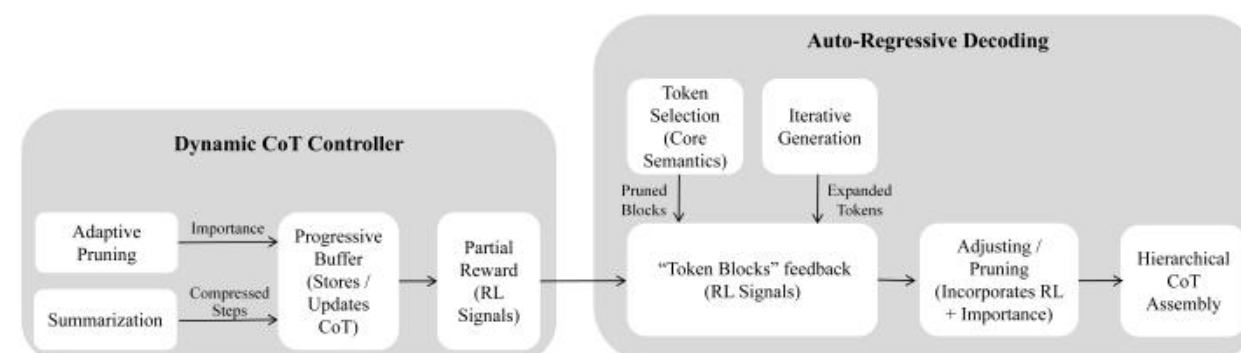

Fig.4 - The connection between dynamic CoT controller and auto-regressive decoding with feedback loop.

*2.2.4 Auto-Regressive Decoding with Feedback Loop (Iterative Small Blocks)*

Auto-regressive decoding with feedback loop calculates the relative contribution of tokens in the current reasoning process through the partial reward estimator. It filters qualified tags into iterative generation and gradually builds a complete reasoning chain. The generation process uses small block decoding to ensure that reasoning can optimize the structure under short-term feedback constraints to improve adaptability and efficiency. Then, adjusting/pruning dynamically selects whether to compress, retain or expand tags based on the partial reward signal and internal dynamic threshold to reduce redundant calculations and ensure efficient reasoning. Some of the filtered tokens are fed back to the progressive reasoning buffer to ensure consistent reasoning logic, and weight estimation is dynamically updated. The mathematical expression of attention merging is:

- Hierarchical Decoding and Expansion Rule

$C_{t+1}$ = $C_t$ ∪ $T_{t+1}$, $T_{t+1}$ = Adjust(Decode ($T_t$, $\Theta_t$, $r_t$), Importance($T_{t+1}$), $r_{t+1}$)

- Reward-Guided Hierarchical CoT Refinement

F(t+1) = Refine (AssembleSplit ($T_{t+1}$), RewardMap ($M_{t+1}$), $r_{t+1}$))

where $C_t$ is CoT structure at time step $t$; $C_{t+1}$ represents update CoT after processing new tokens; $T_t$ is token block at time step $t$; $T_{t+1}$ is adjusted token block after pruning/expansion; $\Theta_t$ is Local model parameters, including gating logs and RL signals; $r_t$ represents partial reward signal guiding RL; $r_{t+1}$ represents partial reward signal guiding hierarchical selection; Importance ($T_{t+1}$) represents function computing token importance; Adjust() is the operator for pruning/expanding token sequences; Decode() represents function generating token sequences from parameters; $F_{t+1}$ is final hierarchical CoT representation at step $t$+1; $M_{t+1}$ is macro-level summary tokens, $m_{t+1}$ is micro-level detail tokens; AssembleSplit() represents operator splitting token blocks into macro/micro segments; RewardMap() represents function ranking macro segments based on reward signals; Refine() represents operator merging macro, micro, and reward-driven structures into final hierarchical CoT.

The filtered and adjusted marker sequences enter the hierarchical CoT assembly, which integrates multi-level information to enhance the accuracy and adaptability of dynamic reasoning (Fig. 5).

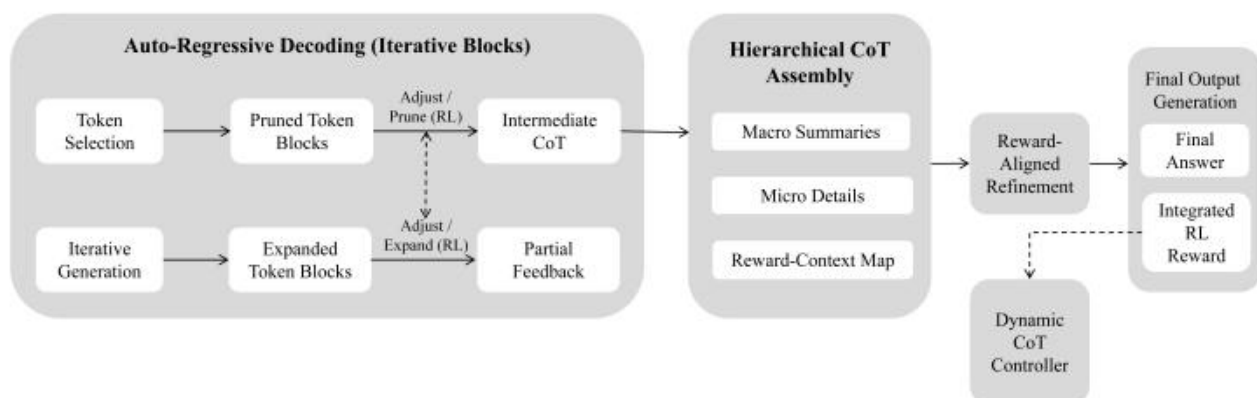

Fig.5 - The connection between auto-regressive decoding with feedback loop and hierarchical CoT assembly.

### *2.2.5 Hierarchical CoT Assembly*

Hierarchical CoT assembly extracts high-level semantic information through macro summary builder and establishes a global reasoning structure to ensure contextual consistency. The fine-grained information is stored through micro detail buffer, which retains important reasoning details to avoid semantic loss. The stored information is further analyzed by the contextual mapper, which enables the reasoning process to dynamically adapt to contextual changes by integrating different levels of information. The reward-aligned refinement adjusts the reasoning weight through the RL mechanism and dynamically adjusts the contribution of key markers based on the learning reward signal. The algorithm is as follows:

- Macro / Micro Separation

$$C_{\text{macro}}, C_{\text{micro}} = \text{Assemble}\ (C, \text{PartialRewards})$$

- Reward-Aligned Refinement

$$C_{\text{final}} = \text{Refine}\ (C_{\text{macro}}, C_{\text{micro}}, \text{RewardMap})$$

where $C$ still represents CoT; $C_{\text{macro}}$ represents is a macro-level reasoning component; $C_{\text{micro}}$ is micro-level reasoning component; Assemble $(\cdot)$ is assembly function, it is responsible for dividing the reasoning steps and decomposing the complete CoT structure into $C_{\text{macro}}$ and $C_{\text{micro}}$; $C_{\text{final}}$ represents final refined CoT; Refine $(\cdot)$ represents refinement function.

### *2.2.6 Final Output Generation*

After completing the hierarchical CoT assembly, the final output generation produces the final answer in response. At the same time, integrated RL reward re-evaluates reasoning weights and adjusts pruning and gating policies through importance estimators, and feeds the updated decision signals back to the dynamic CoT controller to optimize future reasoning processes. The supported algorithms are as follows:

- Final Answer

$$\mathbf{A} = \text{OutputAnswer}(C_{\text{final}})$$

- Compute Episode Reward

$$R_{\text{episode}} = \text{RewardFunction}\ (\mathbf{A}, \text{EnvironmentState})$$

- Update Dynamic CoT Parameters (Loop)

$$\Theta_{\text{CoT}} \leftarrow \Theta_{\text{CoT}} + \eta \nabla_{\Theta\text{CoT}} \cdot R_{\text{episode}}$$

where **A** is final answer; $C_{\text{final}}$ represents the output result from hierarchical CoT assembly; $R_{\text{episode}}$ represents episode reward; $\Theta_{\text{CoT}}$ is is a parameter of D-CoT; $\nabla_{\Theta\text{CoT}} R_{\text{episode}}$ is gradient of episode reward; $\eta$ is the learning rate that controls the update amplitude of the D-CoT parameters.

## III. EXPERIMENTS

D-CoT is essentially a modular enhancement technology for deep reasoning. Its core mechanism includes adaptive step adjustment and reasoning pruning, but does not involve changes to the pre-training architecture of LLMs. This means that it needs to be embedded into existing LLMs to achieve performance as it relies on its language understanding capabilities rather than independent execution (Kaur et al., 2024). Therefore, this research chose simulation experiment to test D-CoT's adaptive step adjustment, computational resource allocation, reasoning pruning and reward alignment in the deep reasoning process. It allows verifying the impact of different regulation strategies on reasoning performance in a controlled environment, which avoids being limited by the immutability of the internal architecture (Edmonds & Hales, 2005; Kleijnen, 2018).

In addition, except for a few open sources such as DeepSeek R1 and MemGPT, mainstream ones such as OpenAI o1 and o3 min-high are still closed-source LLMs, and the API interface cannot provide internal control capabilities for the reasoning steps (Lu et al., 2024). The researcher was unable to directly modify the internal reasoning architecture of LLMs by integrating D-CoT for dynamic step adjustment and reward alignment testing (Arrieta et al., 2025).

### *3.1. Experimental Setup*

Considering that it is difficult to directly integrate into existing closed-source LLMs and a simulation experiment was used to test its reasoning performance, the researcher simulated the process of integrating D-CoT into current LLMs through custom GPTs with runnable code. By developing code in Python 3.13 IDLE and uploading it to the Python simulator based on GPTs, the researcher simulate the dynamic control of D-CoT's deep reasoning steps, calculation mechanism adjustment and reward alignment strategies, and simulate its time application scenarios in LLMs. The Python simulator has the highly complex ability to execute code in customized GPTs, which earned it a 4.2-star rating, ensuring flexibility and controllability in testing (Fig. 6). The researcher uploaded the D-CoT code to a Python-based simulator as an experimental group, and selected DeepSeek R1 with deep reasoning capabilities as a control group to compare the performance differences of reasoning.simulator

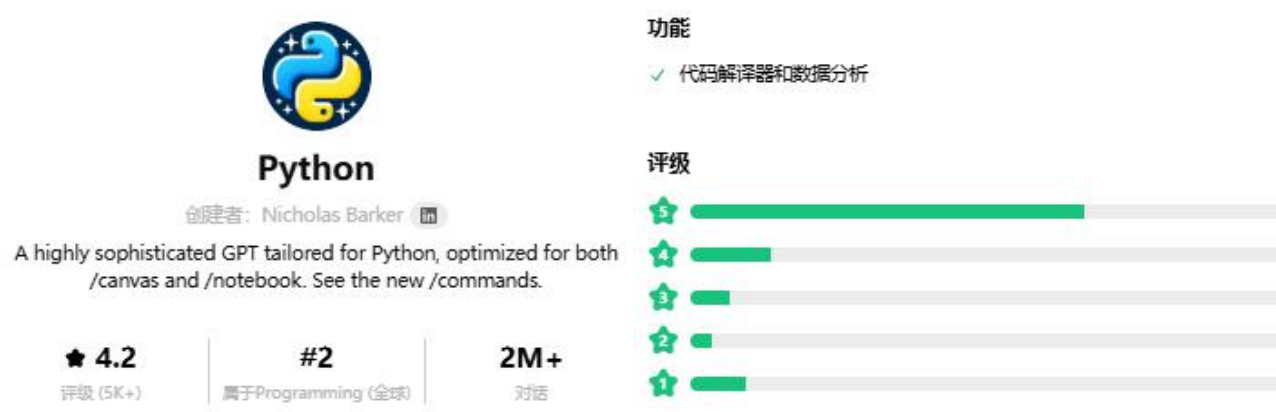

Fig. 6 - Python .simulator based on custom GPTs.

*3.2. Dataset*

Based on the fact that solving linear algebra requires high structure and strong reasoning requirements, it has become an effective test set to evaluate the reasoning ability of CoT. The researcher chose the test questions of 18.06 Spring 2022 Problem Sets and Exams of 18.06 Linear Algebra of MIT OpenCourseWare as experimental data. Linear algebra usually includes multi-step calculations and logical derivation. Both Long CoT and D-CoT need to show the details in the reasoning process. In addition, the test questions cover questions of different difficulty levels and can test the adaptability and robustness of dynamic D-CoT under various reasoning challenges. And because this test question is provided in text format, it is suitable for uploading to GPTs that emulate Python and ensures the independent operation of D-CoT. This data has been placed in Appendix 1, and it is noteworthy that the MIT OpenCourseWare license authorizes the test questions to be used publicly and experimentally for non-commercial purposes.

*3.3. Implementation*

The researcher uploaded the D-CoT code developed by Python 3.13 IDLE to the Python simulator based on GPTs to simulate its integration process in LLMs. At the same time, the researcher also set up corresponding instructions in DeepSeek R1 to execute and process linear algebra test questions. Afterwards, the researcher uploaded the 18.06 Spring 2022 Problem Sets and Exams one by one to the simulator and DeepSeek R1 integrated with D-CoT and recorded the experimental results. In order to avoid deviations in image semantics recognition between the experimental group and the control group that would affect the experimental results, all test questions were converted to machine-readable format and renumbered into 18 questions without changing the original meaning.

In view of the problem of this research, the researcher sets three core evaluation indicators to evaluate the reasoning cost and computing resource consumption of D-CoT. These metrics are reasoning time, CoT length (reasoning steps) and token count to quantify the computational overhead of the D-CoT simulator and DeepSeek R1 respectively. In addition, since this research is committed to reducing computational redundancy and optimizing reasoning steps rather than verifying the accuracy of calculation results, it does not include accuracy score as an evaluation indicator. Complete experimental records and codes have been uploaded to GitHub repository to ensure reproducibility.

## IV. Result & Discussion

After the above execution process, the experimental group and the control group respectively display the data results except for reasoning time, CoT length (reasoning steps) and token count (Fig. 7).

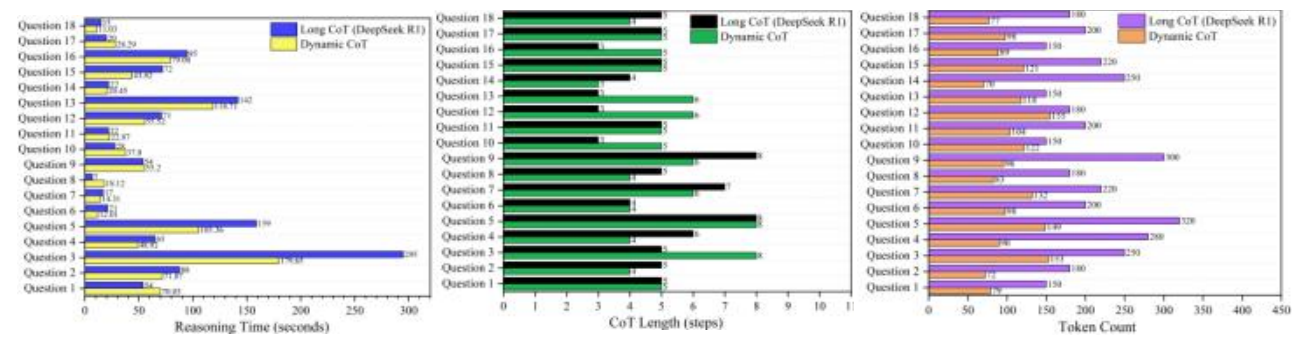

Fig. 7 - Comparison between D-CoT simulator and DeepSeek R1

According to the data results, in terms of reasoning time, the maximum reasoning time of DeepSeek R1 reaches 295 seconds, while D-CoT is significantly reduced to 179.65 seconds, showing its computing optimization capabilities in long reasoning processes. Judging from the distribution trend, the reasoning time of DeepSeek R1 is mostly concentrated in 50 to 150 seconds, but some questions exceed 200 seconds, showing the instability of its reasoning chain length; in comparison, D-CoT mostly maintains in the range of 40 to 120 seconds, and the fluctuation is small, indicating that it can effectively control the reasoning time. In terms of CoT length (reasoning steps), the number of steps of DeepSeek R1 is up to 8 steps, while D-CoT is controlled within 6 steps, and most questions only require 3 to 6 steps, which significantly shows that it can reduce redundant reasoning steps through a dynamic adjustment mechanism. As for the token count, the token usage of DeepSeek R1 is up to 320 tokens, while that of D-CoT is significantly reduced to 180, and its overall distribution is mainly concentrated in the 70 to 180 range. Compared with DeepSeek R1, which exceeded 300 in some complex questions, D-CoT has demonstrated effective suppression of token growth and reduction of computing resource consumption. The results provide the evidence that D-CoT shows significant advantages in reasoning time, CoT length (reasoning steps) and token count usage. It can reduce computational redundancy, optimize the deep reasoning process, and maintain stability in difficult reasoning tasks.

## V. Limitation & Future Research

In light of the fact that external researcher do not have the authority to access and modify the internal architecture of most current mainstream closed-source LLMs, D-CoT is difficult to directly integrate and uses simulation experiments to compare with DeepSeek R1 through a customized GPTs platform. However, the computing power, neural network size and parameters of GPTs are significantly different from DeepSeek R1, which causes the generalizability of the data results to be affected by additional factors. In addition, D-CoT relies on auto-regressive decoding and feedback mechanisms to adjust dynamic reasoning steps. This mechanism causes certain interference in fully reproducing the effects of pruning and reward alignment of LLMs under different contextual conditions in a simulation environment. The above limitations show that obtaining LLMs of open source architecture for internal integration testing in the future will help to more accurately evaluate the actual application performance of D-CoT.

## VI. Conclusion

In view of the fact that long CoT has a large number of intermediate steps in reasoning, which increases reasoning costs and consumption of computing resources due to inherent computational redundancy and delayed feedback, this research proposes dynamic chain-of-thought (D-CoT) to implement a dynamic deep reasoning mechanism of adaptive pruning, reward alignment, and step-wise control. It simulated the integration of D-CoT through simulation experiment in Python 3.13 IDLE combined with the Python simulator based on GPTs, and used DeepSeek R1 based on long CoT as a control group to test its performance in solving the 18.06 linear algebra test questions of MIT OpenCourseWare. Experimental results show that D-CoT can effectively reduce computing resource consumption in reasoning time, CoT length and token count, thereby optimizing and reducing the computing cost and resource consumption of deep reasoning. Although D-CoT is limited by the current closed-source environment and cannot be directly integrated into LLMs for testing, it provides a new feasibility perspective for deep reasoning optimization of LLMs.

## APPENDIX 1

The data for this research comes from the 18.06 Spring 2022 Problem Sets and Exams questions of MIT OpenCourseWare's 18.06 Linear Algebra, which were uploaded to the D-CoT simulator and DeepSeek R1 to answer respectively. In order to avoid the image semantic recognition bias of matrix operations from interfering with the experimental results, the researcher converted all test questions into computer language input and renumbered them into 18 questions without changing the original meaning. This test may be used non-commercially for the experimental purposes of this study under the terms of the Creative Commons Attribution-Non Commercial sharealike (CCBY-NC-SA) license.

**Problem 1** (4+4+6 points):

The matrix A is given by

A = LUL-1U-1

for

$$L = \begin{pmatrix} 1 & & & \\ -1 & 1 & & \\ 0 & 3 & 1 & \\ 1 & 0 & 0 & 1 \end{pmatrix}, U = \begin{pmatrix} 2 & 0 & 1 & 1 \\ & -1 & 0 & -1 \\ & & -2 & 1 \\ & & & 1 \end{pmatrix}.$$

(a) Write an expression for A-1 in terms of L, U , L-1, and/or U-1 (but you don't need to actually multiply or invert the terms!).

(b) What is the determinant of A?

(c) Solve PAx = b for x, where P is the 4 × 4 permutation that swaps the

1stand 4th elements of a vector, and b = $\begin{pmatrix} -5 \\ 4 \\ 11 \\ -3 \end{pmatrix}$. (You can get partial credit by just outlining a reasonable sequence of steps here that doesn't involve a lot of unnecessary calculation.)

**Problem 2** (4+6 points):

(a) If a and x are vectors in Rn , then aaTx can be computed using either left-to-right as (aaT )xor right-to-left as a(aTx), where the parentheses indicate the order of operations. Roughly count the number of arithmetic operations (additions and multiplications) in these two approaches: say whether each approach scales proportional to n, n2 , n3 , etcetera.

(b) A is an n × n real matrix and x is an n-component real vector. Indicate which of the following must be equal to one another:

trace($Axx^T$ ), trace($xAx^T$ ), trace($x^TAx$), $x^TAx$,

trace($x^TxA$), $xx^TA$, trace($xx^TA$), determinant($xx^TA$).

For the expressions that are equal, indicate how you would evaluate this quantity in a cost (in arithmetic operations) proportional to $n^2$.

**Problem 3** (4+4+4+5 points):

You have a 4 × 3 matrix A = ( $q_1$ $2q_2$ $3q_1 + 4q_2$), where we have expressed the three columns of A in terms of the orthonormal vectors

$$q_1 = \frac{1}{2}\begin{pmatrix} 1 \\ 1 \\ -1 \\ -1 \end{pmatrix}, \qquad q_2 = \frac{1}{2}\begin{pmatrix} 1 \\ -1 \\ -1 \\ 1 \end{pmatrix}.$$

(a) What is the rank of A?

(b) Give a basis for N(A).

(c) You are asked to calculate the projection matrix P onto C(A). Your friend Harvey Ard suggests applying the formula P = A(AT A)-1AT he memorized in linear algebra. Explain why this won't work here, and give an even simpler (correct) formula for P in terms of the quantities above. (You need not evaluate P numerically, just write a formula in terms of products of quantities defined above.)

(d) Find the closest vector to x = $\begin{pmatrix} 2 \\ 0 \\ 0 \\ 0 \end{pmatrix}$ in $N(A^T)$.

**Problem 4** (3+4+4+6 points):

The nullspace N (A) of the real matrix A is spanned by the vector v =

$$\begin{pmatrix} 1 \\ 2 \\ 3 \\ 4 \end{pmatrix}$$

(a) Give as much true information as possible about the size (the number of rows and columns) of *A*.

(b) Give an eigenvector and eigenvalue of the matrix B = $(3I - A^TA)(3I + A^TA)^{-1}$ .

(c) Aside from the eigenvalue identified in the previous part, all other eigen- values $\lambda$ of $B$ must be (**circle/copy all that apply**): purely real, purely imaginary, zero, negative real part, positive real part, jλj < 1, jλj > 1, jλj ≤ 1, jλj ≥ 1.

(d) Give a good approximate formula for of $B^n$

$$\begin{pmatrix} 0 \\ -1 \\ 0 \\ 8 \end{pmatrix}$$

for large n. (Give an explicit numerical vector, possibly including simple functions of n like $2^n$ or $n^3$... no other abstract symbolic formulas.)

**Problem 5** (10 points):

Describe (give an explicit numerical result with as few unknowns as possible) all possible linear combinations of the vectors

$$a_1 = \begin{pmatrix} 1 \\ 0 \\ 2 \end{pmatrix},\ a_2 = \begin{pmatrix} 1 \\ 2 \\ 4 \end{pmatrix},\ a_3 = \begin{pmatrix} 1 \\ -1 \\ 3 \end{pmatrix},\ a_4 = \begin{pmatrix} 1 \\ 1 \\ 1 \end{pmatrix}$$

that give the vector $x = \begin{pmatrix} 4 \\ -1 \\ 5 \end{pmatrix}$

**Problem 6** (8+8 points)**:**

Professor May Trix is trying to construct an 18.06 homework question in which $dx/dt = Ax$ has the solution

$$x(t) = v_1 \cos(2t) + v_2 e^{-t} + v_3 \sin(2t)$$

for some **nonzero real** constant vectors $v_1, v_2, v_3$, and some initial condition $x(0)$. Help May construct A, $v_1$, $v_2$, $v_3$, and $x(0)$:

(a) Write down a numerical formula for a possible **real** matrix A such that A is as **small in size as possible** and where A contains **no zero entries**. Your formula can be left as a product of some matrices and/or matrix inverses - you don't need to multiply them out or invert any matrices, but you should give possible numeric values for all of the matrices in your formula. (You don't need to explicitly check that your A has no zero en- tries as long as zero entries seem unlikely. e.g. the inverse of a matrix with no special structure probably has no zero entries.)

(Note that there are many possible answers here, but they will all have certain things in common.)

(b) Using the numbers that you chose from the formula in your previous part, give possible corresponding (numeric) values for $x(0)$, $v_1$, $v_2$, and $v_3$.

## ACKNOWLEDGMENT

The author is grateful to DeepSeek for its contribution in open-source LLMs for providing key technical reference and motivation for this research, which inspired the author to deeply explore the issues of computational redundancy and cost consumption of long CoT. In particular, DeepSeek R1's excellent performance in long-step reasoning and context extension provides important benchmark data for this research. In addition, the technology sharing of the DeepSeek open source community further contributed to the proposal and verification of dynamic CoT. Hereby, the author would like to express his most sincere gratitude to the DeepSeek team and open source contributors.

## CODE AVAILABILITY STATEMENT

This research is also called open source spirit, and all relevant codes are open for use. But it requires users to mark the source and originality. The link to the Github repository of the code is as follows:

https://github.com/brucewang123456789/GeniusTrail/tree/main/Dynamic%20CoT